%% file: main.tex
\documentclass[conference,9pt]{IEEEtran}
\IEEEoverridecommandlockouts

\usepackage{amsmath,amssymb,amsfonts,bm}
\usepackage[colorlinks, linkcolor=magenta, anchorcolor=green, citecolor=blue]{hyperref}
\usepackage{url}
\usepackage{graphicx}
\usepackage{textcomp}
\usepackage{utfsym}
\usepackage{xcolor}
\usepackage{threeparttable}
\usepackage{tabularx}
\usepackage{booktabs}
\usepackage{multirow}
\usepackage{colortbl}
\usepackage{subfigure}
\usepackage{pifont}
\usepackage{algorithmic}
\usepackage[linesnumbered,ruled]{algorithm2e}
\newcommand{\hl}{\textcolor{blue}}

\usepackage{booktabs}
\usepackage{graphicx}
\usepackage{bigstrut,multirow,rotating}
\usepackage{booktabs}
\usepackage{multirow}
\usepackage{bigstrut}
\usepackage{graphicx,pifont}
\let\oldding\ding
\renewcommand{\ding}[2][1]{\scalebox{#1}{\oldding{#2}}}
\usepackage{setspace}
\usepackage{mathrsfs}

\usepackage[top=0.7in,bottom=0.70in,left=0.58in,right=0.58in]{geometry}
\iftrue
\fi

\usepackage{amsmath,amssymb,amsfonts}
\usepackage[numbers,sort&compress]{natbib}
\newcommand*{\circled}[1]{\lower.7ex\hbox{\tikz\draw (0pt, 0pt)%
    circle (.5em) node {\makebox[1em][c]{\small #1}};}}

\def\BibTeX{{\rm B\kern-.05em{\sc i\kern-.025em b}\kern-.08em
    T\kern-.1667em\lower.7ex\hbox{E}\kern-.125emX}}
    
\usepackage{authblk}
\begin{document}

\title{
Hardware-Aware Graph Neural Network Automated Design \\ 
for Edge Computing Platforms
\thanks{This work is supported by National Natural Science Foundation of China (Grant No. 62072019). Corresponding authors are Jianlei Yang and Chunming Hu, Email: \url{jianlei@buaa.edu.cn}, \url{hucm@buaa.edu.cn}.
}
}

\author{Ao Zhou, Jianlei Yang, Yingjie Qi, Yumeng Shi, Tong Qiao, Weisheng Zhao, Chunming Hu}

\affil{Beihang University, Beijing, China}


\IEEEoverridecommandlockouts
\IEEEpubid{\makebox[\columnwidth]{
979-8-3503-2348-1/23/\$31.00~\copyright 2023 IEEE
 \hfill}
\hspace{\columnsep}\makebox[\columnwidth]{ }}

\maketitle
\IEEEpubidadjcol

\input{0-abstract.tex}
\input{1-introduction.tex}
\input{2-motivation.tex}
\input{3-framework.tex}
\input{4-experiment.tex}
\input{5-conclusions.tex}

\bibliographystyle{unsrt}

\scriptsize
\begingroup
\bibliography{dac2023}
\endgroup

\end{document}

%% file: 0-abstract.tex
\begin{abstract}

Graph neural networks (GNNs) have emerged as a popular strategy for handling non-Euclidean data due to their state-of-the-art performance.
However, most of the current GNN model designs mainly focus on task accuracy, lacking in considering hardware resources limitation and real-time requirements of edge application scenarios.
Comprehensive profiling of typical GNN models indicates that their execution characteristics are significantly affected across different computing platforms, which demands hardware awareness for efficient GNN designs.
In this work, HGNAS is proposed as the first \underline{H}ardware-aware \underline{G}raph \underline{N}eural \underline{A}rchitecture \underline{S}earch framework targeting resource constraint edge devices.
By decoupling the GNN paradigm, HGNAS constructs a fine-grained design space and leverages an efficient multi-stage search strategy to explore optimal architectures within a few GPU hours.
Moreover, HGNAS achieves hardware awareness during the GNN architecture design by leveraging a hardware performance predictor, which could balance the GNN model accuracy and efficiency corresponding to the characteristics of targeted devices.
Experimental results show that HGNAS can achieve about $10.6\times$ speedup and $88.2\%$ peak memory reduction with a negligible accuracy loss compared to DGCNN on various edge devices, including Nvidia RTX3080, Jetson TX2, Intel i7-8700K and Raspberry Pi 3B+.
 


\end{abstract}

%% file: 1-introduction.tex
\section{Introduction}

Graph neural networks (GNNs) have achieved state-of-the-art performance in various graph representation scenarios, including node classification \cite{kipf2017semi}, link prediction \cite{zhang2018link}, recommendation \cite{ying2018graph} and 3D representation learning on point clouds \cite{qi2017pointnet++}. 
Due to the powerful feature extraction capabilities on topological structures, GNNs have become a popular strategy for handling point cloud data, such as DGCNN \cite{wang2019dynamic}.
Moreover, with the rising popularity of 3D scanning sensors in edge devices, such as mobiles, unmanned aerial vehicles, etc., it is natural to investigate how to design efficient GNN models for edge applications.

\begin{figure}[t]
    \centering
    \includegraphics[width = 1\linewidth]{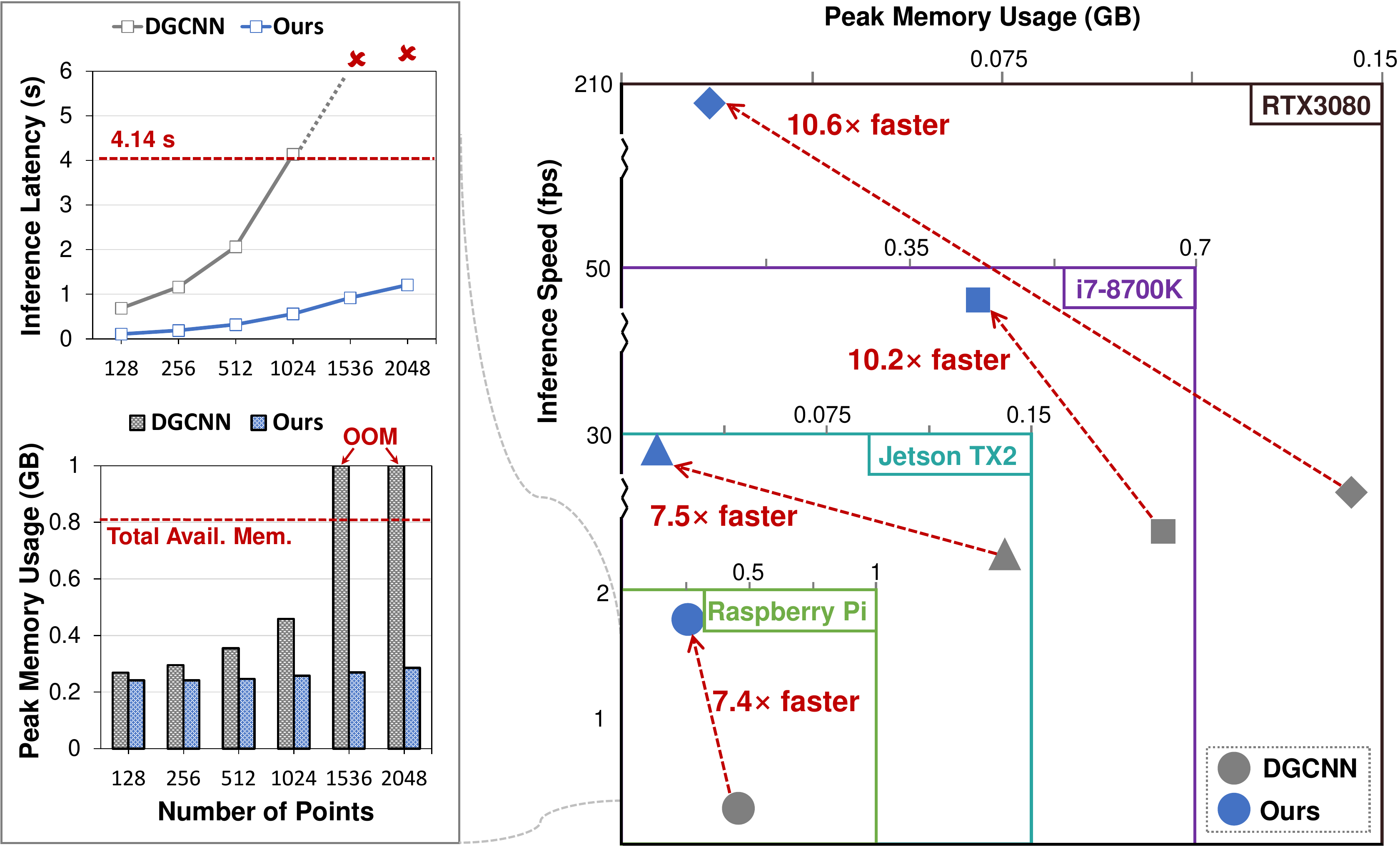}
    \caption{Comparison between our approach and DGCNN. Inference latency and peak memory usage on Raspberry Pi are illustrated on the left. Inference speed and memory efficiency improvement across different edge devices are illustrated on the right.}
    \label{fig:intro-comparison}
    \vspace{-3pt}
\end{figure}

The most challenging issue of GNNs is their hungry demands for hardware resources. 
Especially for resource-constrained edge devices, GNNs usually exhibit low hardware efficiency and \textit{Out-Of-Memory} (OOM) problem, limiting their potential to handle real-world edge applications.
For instance, given a frame of point cloud data, DGCNN performs KNN operations in each GNN layer for graph construction, resulting in poor hardware efficiency on edge platforms.
As shown in Fig. \ref{fig:intro-comparison}, the examined GNN's inference latency and peak memory usage for handling point cloud classification tasks continues increasing rapidly as the number of processed points grows.
Specifically, for the default setting with $1024$ points, DGCNN \cite{wang2019dynamic} needs more than \textbf{4 seconds} to handle a single frame on Raspberry Pi (i.e. about $1/4$ fps, frame per second), which is intolerable for real-time applications.
In addition, processing the involved graphs with more than $1536$ points will cause OOM problems during DGCNN inference.
Therefore, deploying GNNs on edge devices is extremely challenging with resource constraints. 

Several handcrafted approaches have been proposed to tackle the prohibitive inference cost of GNNs in point cloud processing
\cite{li2021towards, tailor2021towards, zhang2021exploration}. 
Even though they could achieve notable speedups by simplifying the GNN computations, manual optimization is difficult to adapt for different computing platforms, regarding the huge design space and various hardware characteristics \cite{tailor2021towards}.
As a very promising approach, hardware-aware neural architecture search (NAS) could explore optimal architectures automatically according to given objectives, which is independent of human experience and avoids manual labor.
By leveraging this approach, several works have made encouraging progress in developing efficient DNN models on edge devices~\cite{dudziak2020brp, wu2019fbnet}.
More specifically, we aim to exploit hardware-aware NAS for efficient GNNs derived from point cloud applications on edge platforms.

Although several NAS frameworks have been introduced for GNNs \cite{ijcai2020p195, gao2022graphnas++}, most of them have rarely considered the hardware constraints and latency requirements inspired by real-world edge applications.
In this paper, HGNAS is proposed as an efficient hardware-aware graph neural architecture search framework for point cloud applications on edge computing platforms. 
Such a hardware awareness is achieved by integrating hardware efficiency as a partial objective when performing the single-path one-shot evolutionary search.
Given the targeted edge devices, HGNAS automatically explores the optimized graph neural architectures by guaranteeing both accuracy and efficiency under hardware constraints (i.e. inference latency, model size, etc.).
One straightforward approach to achieve hardware awareness is to deploy the generated architecture candidates on the targeted devices and feedback the measured data during exploration.
However, the required tremendous real-time hardware measurement leads to very inefficient exploration process since the cost of edge inference and communication is often unbearable.
As such, a more elegant way to evaluate GNN hardware efficiency for different platforms is warranted.
In addition, the redundant operations within the layer-wise GNN design space and lengthy search times also pose great challenges to the efficient GNNs exploration process.

To address these challenges, our proposed HGNAS framework leverages novel hardware-aware techniques for procuring both desirable GNN performance and search efficiency.
We spot that GNN architectures themselves are also graphs that can be well represented by GNNs.
Such a concept is put into good use by drawing on the idea of \textit{Use GNN to perceive GNNs}.
Therefore, HGNAS can effectively perceive the latency of GNN candidates through a well-polished GNN-based predictor. 
Furthermore, we develop a fine-grained design space composed of basic operations to unleash the potential of GNN computations. 
HGNAS also adopts an efficient multi-stage hierarchical search strategy by dividing the GNN design space, reducing the search time to a few GPU hours.
As shown in Fig.~\ref{fig:intro-comparison}, HGNAS has been proven superior in both latency and peak memory usage across various edge devices. 
Specifically, HGNAS on resource-constrained Jetson TX2 could achieve the same level of inference latency compared with DGCNN on powerful RTX3080, providing $47\times$ (i.e. $350$W vs. $7.5$W) power efficiency improvement without accuracy loss.
In summary, the contributions of this work are listed as follows:
\begin{itemize}
    \item \textbf{Framework.} To the best of our knowledge, HGNAS is the first NAS framework to perform efficient graph neural architecture search for resource-constrained edge devices. HGNAS can automatically explore GNN models with multiple objectives (accuracy, latency, etc.) for targeted platforms.  
    \item \textbf{Hardware awareness.} To the best of our knowledge, HGNAS is also the first work to achieve hardware performance awareness for GNNs across edge devices. The proposed GNN-based predictor can perceive the latency of a given GNN architecture on the targeted platform in milliseconds.
    \item \textbf{Evaluation.} We have conducted extensive experiments on point cloud classification tasks. By deploying HGNAS-designed models on various edge devices, we achieve about $10.6\times$ speedup and $88.2\%$ peak memory reduction with a negligible accuracy loss.
\end{itemize}

%% file: 2-motivation.tex
\section{Related Works and Motivation}

\begin{figure}[t]
    \centering
    \includegraphics[width = 1\linewidth]{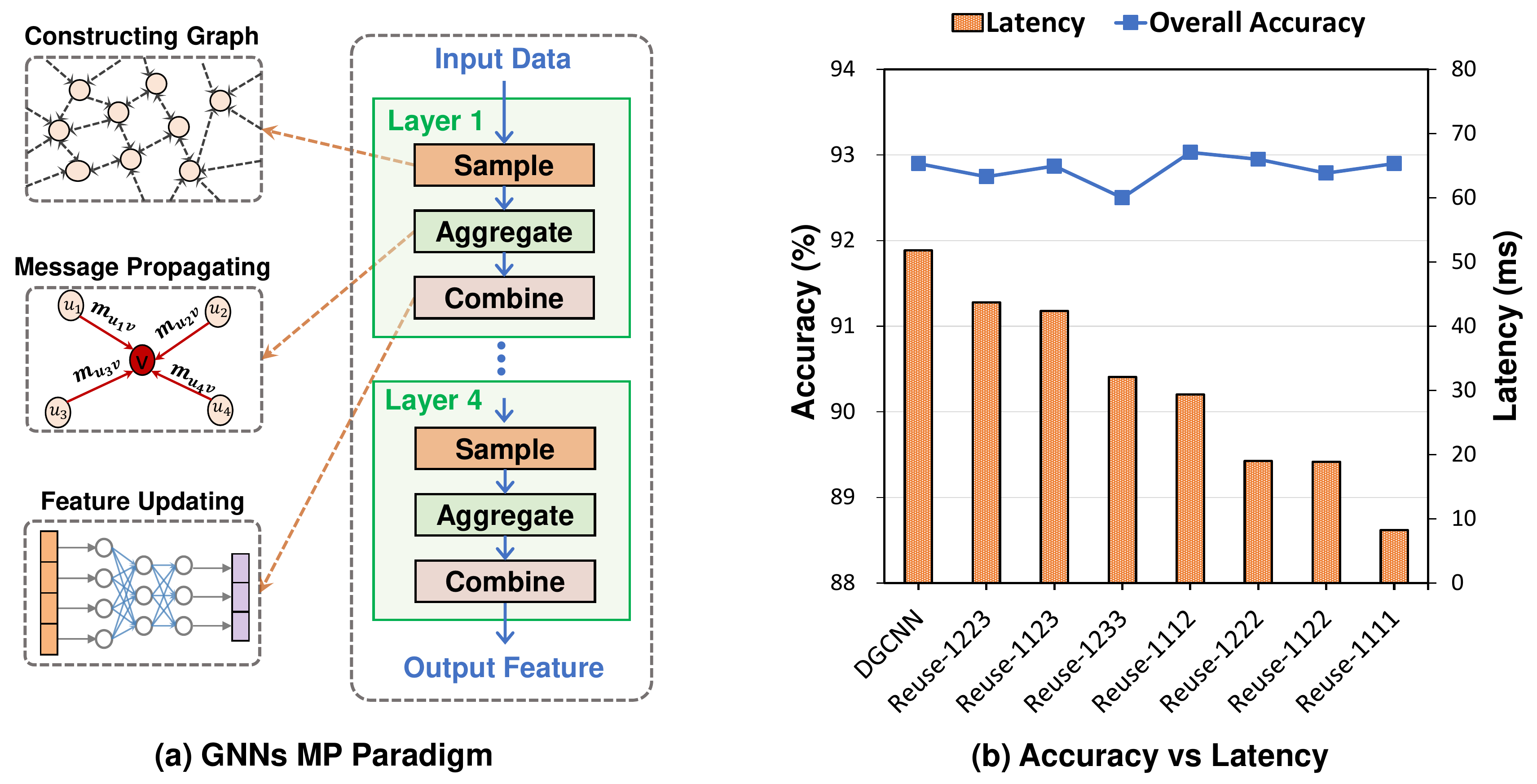}
    \caption{(a) Typical GNN Pipeline with MP paradigm. (b) Accuracy and latency comparison when performing sampled results reuse among different DGCNN layers on ModelNet40~\cite{wu20153d} dataset and RTX3080 platform.}
    \label{fig:motivation-pipeline}
    \vspace{-3pt}
\end{figure}

In combination with previous related works, we introduce several important observations on motivating efficient GNN explorations for edge computing platforms.

\textbf{Observation \raisebox{-0.1em}{\ding[1.3]{172}}}: Redundant operations bring significant overhead.

Generally, GNN computations follow the \textit{message massing} (MP) paradigm, which consists of \textit{graph sampling}, \textit{aggregation}, and \textit{combination}, as shown in Fig.~\ref{fig:motivation-pipeline}(a). 
By reusing operation results across GNN layers, we observe that redundancy may exist within the MP paradigm.
As illustrated in Fig.~\ref{fig:motivation-pipeline}(b), reusing the sampled results between GNN layers does bring some negligible accuracy loss, but could provide a considerable boost to computational efficiency.
It indicates that redundant operations usually bring significant overhead, which is one of the major obstacles in optimizing computational efficiency.

\textbf{Motivation \raisebox{-0.1em}{\ding[1.3]{182}}}: Decouple the MP paradigm by leveraging fine-grained GNN design space.

The above results demonstrate that building GNNs by stacking generic GNN layers together will inevitably bring redundant operations.
The most straightforward approach for this problem is optimizing GNN models through large amounts of ablation studies and analysis. 
Typically some redundant sampling operations are eliminated in DGCNN for achieving better efficiency \cite{li2021towards}.
However, these handcrafted approaches heavily rely on design experience and are non-reproducible for different tasks. 
Inspired by the success of designing scalable GNN models by decoupling GNN paradigm \cite{cai2021rethinking, zhang2022pasca}, GNN layers are decoupled into operations for building a fine-grained design space in an operation-wise manner.
With the freedom offered by fine-grained design space, various configurations (i.e. aggregation range, operation order, etc.) could be generated by learning rather than manual laboring.

\textbf{Observation \raisebox{-0.1em}{\ding[1.3]{173}}}: Exploring fine-grained design space is costly.

Compared with layer-wise design space, the fine-grained design space takes operations as candidate options instead of layers, greatly expanding the scope of exploration.
For example, the backbone of DGCNN consists of four GNN layers, and each layer includes three basic operations, i.e. \textit{sample}, \textit{aggregate}, \textit{combine}. Therefore, to cover most DGCNN architectures, the fine-grained design space must contain at least $12$ positions, $3$ candidate operations, and $N$ functions for each operation (details in Sec.~\ref{sec:design-space}). As a result, there are staggering ${(3N)}^{12}$ options to explore in the design space, which significantly increases the exploration complexity.

\textbf{Motivation \raisebox{-0.1em}{\ding[1.3]{183}}}: Speedup exploration by search space simplification.

The exploration complexity demands a very efficient search strategy to navigate through the huge options.
Intuitively, some studies \cite{zhang2020autoshrink} have proposed to shrink the design space for better search efficiency.
Unfortunately, the boost in efficiency comes with the price of potentially discarding some valued choices.
In reality, the contribution of different layers towards the overall accuracy within a model may vary greatly \cite{sheng2022larger}.
The representational power of front layers usually contributes more to the whole GNNs \cite{tailor2021towards}.
Hence, they simplify the latter parts of DGCNN to obtain better model performance. 
All of these inspire us to improve exploration efficiency by dividing the search space and guiding the search tendencies at different positions.

\begin{figure}[t]
    \centering
    \includegraphics[width = 1\linewidth]{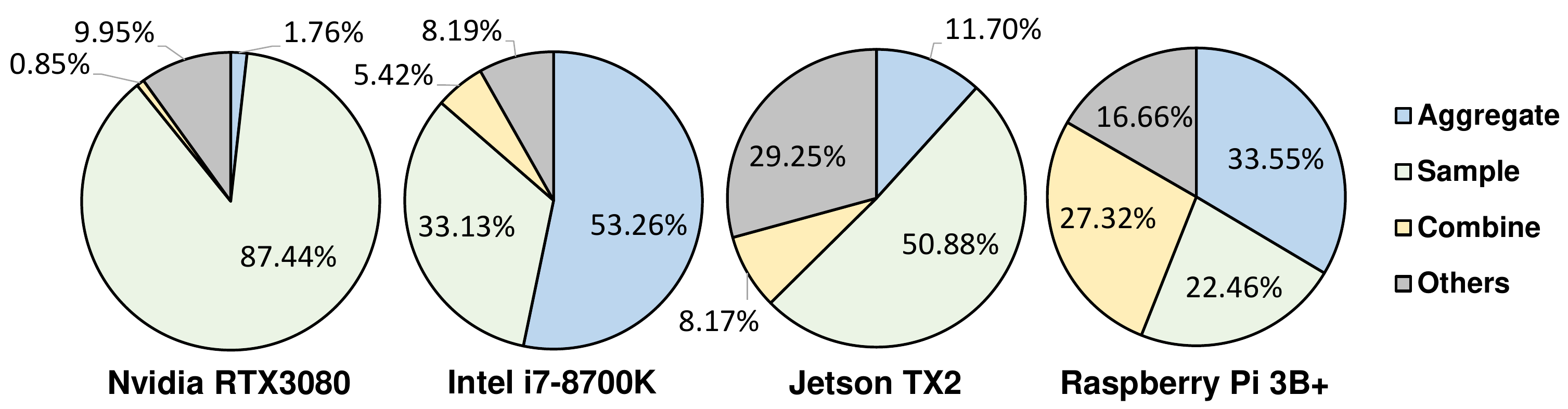}
    \caption{Execution time breakdown of DGCNN across different platforms.}
    \label{fig:motivation-breakdown}
    \vspace{-3pt}
\end{figure}

\begin{figure*}[t]
    \centering
    \includegraphics[width = 1\linewidth]{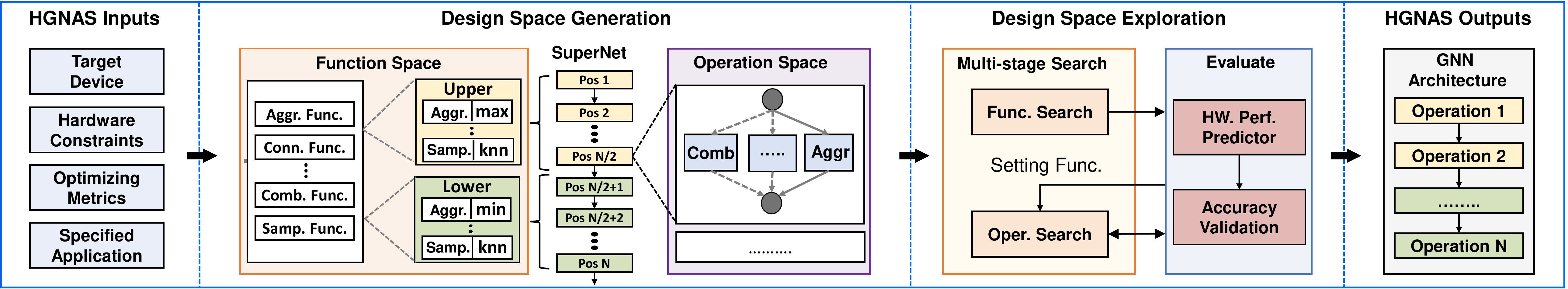}
    \caption{Overview of the HGNAS framework. Oper. and Func. denotes Operation and Function respectively.}
    \label{fig:framework-overview}
    \vspace{-3pt}
\end{figure*}

\textbf{Observation \raisebox{-0.1em}{\ding[1.3]{174}}}: The same GNN model may behave differently on various computing platforms.

Detailed execution time breakdown of DGCNN is illustrated in Fig.~\ref{fig:motivation-breakdown} for different platforms.
These results are obtained by PyTorch Profiler~\cite{profiler}. 
For Nvidia RTX3080 and Jetson TX2, the \textit{sample} operation occupies the majority of execution time. 
This is because GPUs are better at handling compute-intensive matrix operations, and not so good at memory-intensive graph sampling operations.
For Intel i7-8700K, \textit{aggregate} and \textit{sample} take up most of the execution time, which is caused by a massive number of irregular memory accesses.
As the results indicate, the execution of DGCNN on the above platforms belongs to \texttt{I/O-bound}.
However, on the Raspberry Pi, due to the limited resources, all three phases occupy relatively large proportions of execution time, which makes the execution process also \texttt{compute-bound}.
They demonstrate that the same GNN model may result in various execution characteristics across computing platforms. 

\textbf{Motivation \raisebox{-0.1em}{\ding[1.3]{184}}}: Perceive the hardware sensitivities of GNNs across different computing platforms.

Due to differences in hardware architectures and available resources, GNNs that are computationally efficient on GPUs may not be sufficient on other platforms. 
Moreover, the hybrid execution mode of GNNs, which consists of both memory-intensive and computation-intensive operations, poses great challenges for effectively perceiving GNNs hardware sensitivities~\cite{yan2020characterizing}.
Some works attempt to improve hardware efficiency by analytical estimation for hardware/algorithm co-design~\cite{zhang2021g}. 
However, these approaches introduce obvious accuracy drop and are difficult to extend to other platforms~\cite{benmeziane2021comprehensive}.
Although real-time measurement could really capture hardware behaviors, its tremendous overhead is intolerable for efficient exploration.
Therefore, an efficient and scalable hardware-aware approach is highly demanded for GNNs exploration.
Inspired by \cite{dudziak2020brp}, a GNN-based end-to-end hardware performance predictor is integrated to efficiently and accurately perceive GNNs hardware performance across various platforms.

%% file: 3-framework.tex
\section{HGNAS Framework}

\subsection{Problem Definition and HGNAS Overview}

In this paper, we aim to co-optimize the accuracy and hardware efficiency of GNNs on edge devices.
Given a target edge device $\cal{H}$, and hardware constraints $\cal C$, the multi-objective optimization process of HGNAS can be
formulated as: 
\begin{equation}
 \arg \mathop {\max }\limits_{\{{\cal A}, \cal{H}\} } \left( \alpha  * {acc_{val}}\left({{\cal W}^ * }, {\cal A} \right) - \beta  * lat \left({\cal A}, \cal{H} \right) \right) ,
    \label{equ1}
\end{equation}
\begin{equation}
    \begin{split}
            s.t. \quad & {{\cal W} ^ * } = \arg \mathop {\max }\limits_{\mathcal W}{acc_{train}}({\cal W}, {\cal A}) \\
            & lat ({\cal A}, \cal{H}) < {\cal C}
    \end{split},
    \label{equ2}
\end{equation}
where $\cal W$ denotes the model weights, $acc_{train}$ is the training accuracy, $acc_{val}$ is the validation accuracy, $lat$ is the inference latency on targeted platform $\cal{H}$, $\cal A$ is the GNN architecture candidate, $\alpha$ and $\beta$ are the scaling factors used to adjust the optimize propensity between accuracy and latency.

Fig.~\ref{fig:framework-overview} shows the overview of our HGNAS framework. 
Given a specific task, a target device, the hardware constraints, and the optimizing metrics, HGNAS will first generate a \textit{fine-grained operation-based design space} which consists of the \textit{Function Space} and \textit{Operation Space}.  
HGNAS then constructs a supernet covering the GNN design space to adopt the one-shot exploration approach.
Afterward, HGNAS will explore the hierarchical design spaces based on the proposed \textit{multi-stage hierarchical search strategy}. 
During exploration, the evaluation result of each candidate architecture is determined by both the accuracy on the validation dataset, and the hardware performance on the target device.
The hardware performance of GNNs is provided by the \textit{GNN hardware performance predictor} integrated in HGNAS.
In subsequent sections, we will detail the three main components of HGNAS.

\subsection{Fine-grained Operation-based Design Space}\label{sec:design-space}

\textbf{The GNN supernet.}
To lift the restrictions of the traditional GNN design space, instead of presetting the number of GNN layers, HGNAS builds the design space upon positions for GNN operations to achieve more flexibility.
Specifically, for each position in the supernet, there are four basic operations including \textit{connect}, \textit{aggregate}, \textit{combine}, and \textit{sample}, each containing specific attributes.
Aside from the operations derived from the MP paradigm, the \textit{connect} operation, including direct connection and skip-connection, provides more freedom in GNN model construction.
For point cloud processing applications, the message type attribute of aggregate operations specifies the construction method of the messages to be aggregated.
As shown in Fig.~\ref{fig:framework-overview}, the supernet is organized by stacking all the positions together. 
In practice, supernet training demands that operations within each position must obtain the same hidden dimension length.
HGNAS appends linear transformations to operations incapable of altering hidden dimensions, such as \textit{sample} and \textit{aggregate}, to ensure dimension alignment among operations.
These linear transformations will be disposed of in the finalized architecture to avoid introducing additional overhead.

\begin{table}[t]
  \small
  \centering
  \renewcommand\arraystretch{1.3}
  \caption{The available choices in GNN's supernet.}
    \begin{tabular}{l|l}
    \hline
    \textbf{Operation} & \textbf{Function} \\
    \hline
    Connect & Skip-connect, Identity \\
    \hline
    \multirow{2}[4]{*}{Aggregate} & Aggregator type: sum,  min,  max, mean \\
    \cline{2-2}   & Message type: Source pos, Target pos, Rel pos, \\ & Distance, Source$||$Rel pos, Target$||$Rel pos, Full \\
    \hline
    Combine & $8$, $16$, $32$, $64$, $128$, $256$ \\
    \hline
    Sample & KNN, Random \\
    \hline
    \end{tabular}
  \label{tab:space}
  \vspace{-3pt}
\end{table}

\textbf{The hierarchical design space.}
HGNAS detaches attributes from operations to construct an independent \textit{Function Space} to further decouple the fine-grained design space.
The remainder of design space made up of operation types is then used to assemble the \textit{Operation Space}.
In addition, these two sub-spaces can be explored separately by leveraging the proposed multi-stage hierarchical search strategy (see Sec.~\ref{sec:strategy}) to reduce exploration complexity.
All candidate operations and functions are listed in Tab. \ref{tab:space}.

\subsection{Multi-stage Hierarchical Search Strategy}\label{sec:strategy}

HGNAS divides the search process into two stages corresponding to \textit{Function Space} and \textit{Operation Space}, in order to reduce the exploration complexity of the fine-grained design space as aforementioned in \textbf{Observation \raisebox{-0.1em}{\ding[1.3]{173}}}.
Inspired by \cite{guo2020single}, the search strategy in HGNAS is based on an evolutionary algorithm ($EA$), as illustrated in Alg. \ref{gnnalgo}.
In particular, HGNAS first searches a set of functions for GNN supernet from function space.
After the optimal function set is determined, HGNAS trains the GNN supernet and performs a multi-objective operation search for all the positions.
Finally, the algorithm outputs the top-performing model within the design space.
During supernet training, a random sub-network is generated by sampling operations for each position in the GNN supernet, whose weights $\mathcal W$ are updated via back propagation.
Details of the multi-stage search strategy are presented as follows.

\textbf{Stage 1: Function Search.} 
In this stage, HGNAS aims to find a function setting to maximize the supernet accuracy.
To further improve the exploration efficiency, HGNAS divides $N$ positions in GNN supernet into two halves, and shares a set of functions among the \textit{Upper} half ($0,...,N/2$) and another set among the \textit{Lower} half ($N/2+1,...,N$). 
For a supernet with $12$ positions, through sharing functions among positions in the decoupled design space, HGNAS can reduce the number of exploration candidates from $\bm{4.2 \times {10^{12}}}$ to $\bm{1.7 \times {10^7}}$.
Finally, an optimal function set $\mathbb{F}$ is determined for initializing the supernet ${\cal N}_{super}$. 
Note that fixing $\mathbb{F}$ in this stage will significantly reduce the complexity of subsequent operation searches.

\textbf{Stage 2: Operation Search.}
By pre-training the supernet and performing EA-based searches, HGNAS obtains a set of operations that maximizes the objective function value in the \textit{Operation Space}. 
Benefiting from the one-shot search strategy, supernet training and operation search are divorced to avoid the exorbitant cost of sub-network retraining during the search.
To improve the search efficiency and meet the hardware constraints on the targeted device, HGNAS evaluates the candidate architectures based on the proposed hardware performance predictor (see Sec.~\ref{sec:predictor}) during the search.
Only the architectures that meet the hardware constraints will be further evaluated for the accuracy metric.
Specifically, the objective function during the operation search is formulated as:
\begin{equation}
{{\cal{F}}_{obj}(\cal C)} = \left\{ {\begin{array}{*{20}{c}}
{0 \quad \quad \quad \quad \quad \quad \quad \quad ,  \quad if \quad lat \ge \cal C}\\
{\alpha  * ac{c_{val}} - \beta  * lat ,  \quad if \quad lat< \cal C}
\end{array}} \right.
    \label{equ3}
\end{equation}

\begin{algorithm}[t]
\small
\textbf{Inputs:} population size $P$, hardware constraints $\cal C$, target device $\mathcal{H}$, operation space $\mathbb{S}_{op}$, function space $\mathbb{S}_{f}$, max iteration $T$, number of positions $N$.

\textbf{Outputs:} the best found GNN design $\cal A^*$ for target device $\mathcal{H}$.

Initialize GNN supernet ${\cal N}_{super}$ with $N$ positions and two function sets $upper \leftarrow \mathbb{\O}$, $lower \leftarrow \mathbb{\O}$ \\
\hl{/* Stage 1: Function search */} \\

Assign function set: \parbox[t]{0.5\linewidth}{
${\cal N}_{super}[0, N/2] \leftarrow upper$, \\
${\cal N}_{super}[N/2+1, N] \leftarrow lower$}

\For{$1 \leq t \leq T$}{
$\{upper, lower\} \leftarrow EA(P, {\cal N}_{super}, \mathbb{S}_{f}, obj = max(acc_{val}))$\\
}
Fix function set $\mathbb{F}  \leftarrow \{upper,lower\}$ for ${\cal N}_{super}$\\ 
Re-initialize and pre-train ${\cal N}_{super}(\mathbb{S}_{op}, \mathbb{F})$\\
\hl{/* Stage 2: Operation search */} \\
Initialize operation set $\cal O \leftarrow \mathbb{\O} $\\
\For{$1 \leq t \leq T$}{
    ${\cal O} \leftarrow EA({\cal P}, {\cal N}_{super}, \mathbb{F}, S_{op},
    obj= max({\cal F}_{obj}({\cal C})))$\\
}
\textbf{return} optimal architecture ${\cal A^*} \leftarrow \{{\cal O}, \mathbb{F}\}$

\caption{\small Multi-stage hierarchical search strategy.}
\label{gnnalgo}
\end{algorithm}

\subsection{GNN Hardware Performance Predictor}\label{sec:predictor}

\begin{figure}[t]
    \centering
    \includegraphics[width = 1\linewidth]{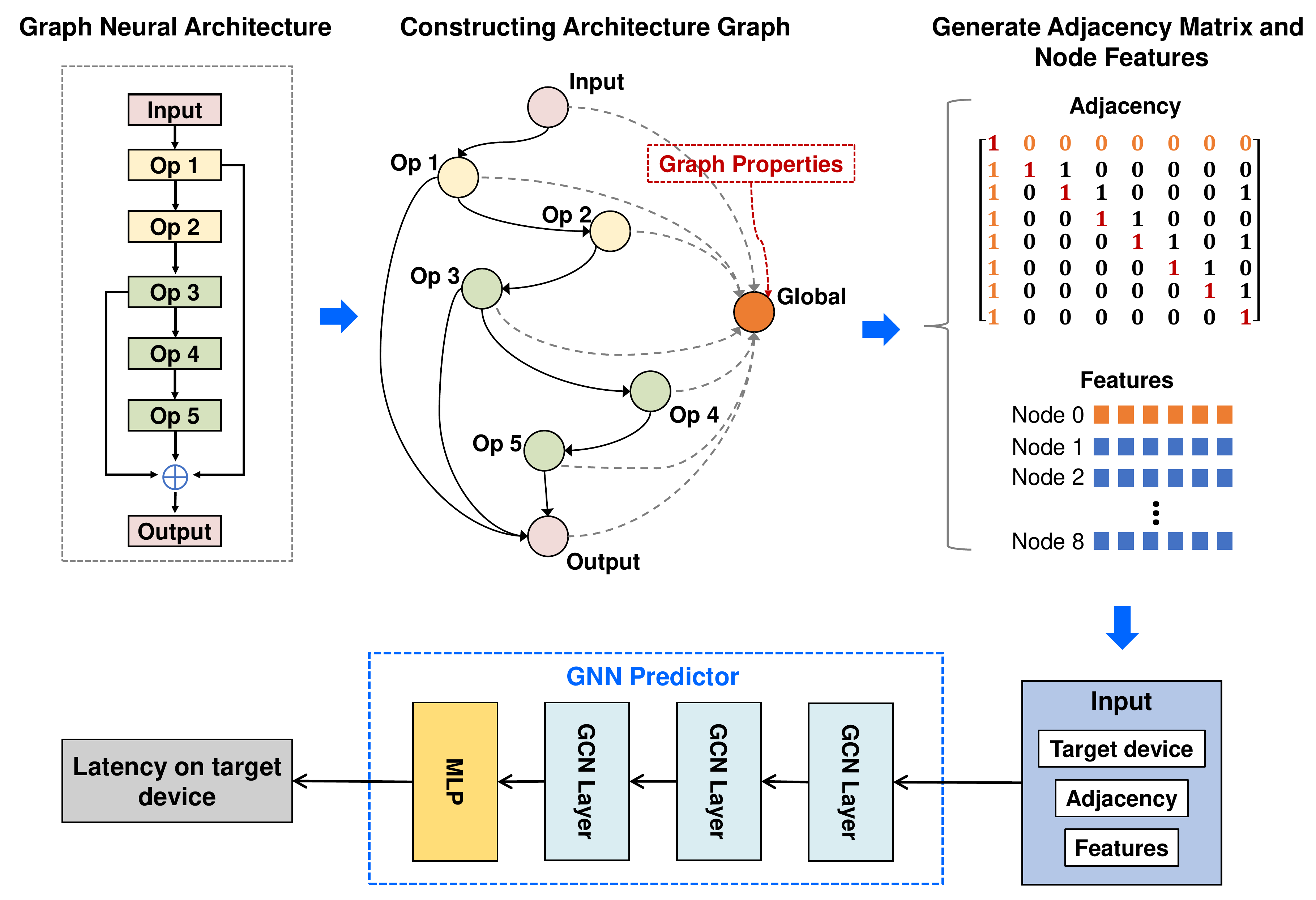}
    \caption{Latency prediction of a candidate model for the target device.}
    \label{fig:predictor}
    \vspace{-3pt}
\end{figure}

To meet the efficiency requirements on targeted devices, we build a GNN hardware performance predictor that can efficiently learn the relationship between GNN architectures and hardware efficiency.
Specifically, its execution process consists of the following phases: \textit{graph construction}, \textit{node feature generation}, and \textit{inference latency prediction}, as shown in Fig.~\ref{fig:predictor}.

\textbf{Graph construction.} 
During this phase, HGNAS abstracts GNN architectures into directed graphs as the input of the GNN predictor.
The nodes in these architecture graphs represent inputs, outputs, and operations, while the edges represent the dataflow within the GNN architecture. 
In practice, accurate prediction of hardware efficiency requires both candidate model architecture and the graph property of the input dataset, which GNN execution highly depends on.
However, the plain abstraction of the original GNN architectures is too sparse for the predictor to obtain enough structural features, while lacking the necessary information on input data.
Hence, HGNAS introduces a global node connected with all nodes in the graph to improve the graph connectivity.
The input data information is also encoded into the global node for better prediction accuracy.

\textbf{Node feature generation.} 
For an operation node, the node feature consists of the operation type and its corresponding function.
Specifically, HGNAS encodes these two components into a $7$-dimensional and a $9$-dimensional one-hot vector respectively, and concatenates the results to represent the node feature.
For input and output nodes, HGNAS assigns them with a zero vector.
For the global node, HGNAS encodes the input graph data properties (number of nodes, density, etc.) into a $16$-dimensional vector as the global node feature.

\textbf{Latency prediction.} 
To avoid the over-smoothing problem often induced by deeper GNNs on small-scale graphs (i.e., the abstracted architecture graph), the predictor consists of only three GCN layers~\cite{kipf2017semi} and a multi-layer perceptron (MLP).
The inputs of the predictor are information on the target device, adjacency matrix, and node features. 
Specifically, the GCN layers utilize the sum aggregator with hidden dimensions of $256 \times 512 \times 512$.
The three layers in MLP with hidden dimensions of $256 \times 128 \times 1$, are followed by a LeakyReLU function for generating a scalar prediction of latency.
As the architecture graphs normally contain no more than a few dozen of nodes, the overhead brought by latency prediction is mostly negligible.
For instance, the GNN-based predictor can predict the latency of a candidate architecture for a target edge device within milliseconds on the Nvidia RTX3080.

%% file: 4-experiment.tex
\section{Experiments}

\begin{figure*}[htb]
    \centering
    \includegraphics[width = 1\linewidth]{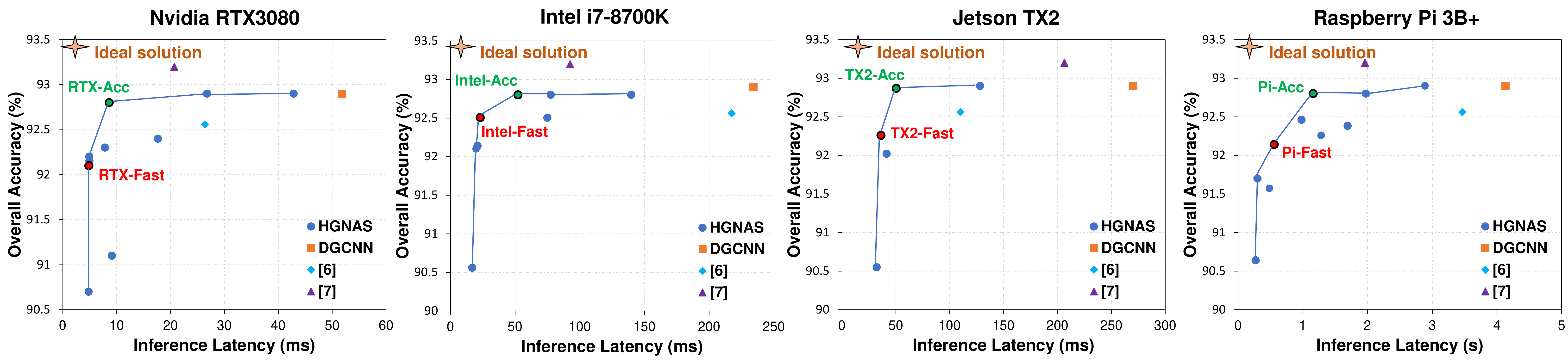}
    \caption{Comparison between existing models and HGNAS across various devices.}
    \label{fig:pareto}
    \vspace{-6pt}
\end{figure*}

\subsection{Experimental Setup}

\textbf{Baselines and datasets.}
For evaluating HGNAS, we consider three baselines: the popular point cloud processing model DGCNN~\cite{wang2019dynamic}, and two manually optimized methods on the DGCNN architecture~\cite{li2021towards, tailor2021towards}.
Our experiments are conducted on the public benchmark ModelNet40~\cite{wu20153d} for classification task with $1024$ points, and following the default hyperparameter settings in~\cite{tailor2021towards}. 
In addition, all the experimental and profiling results are developed on PyTorch Geometric (PyG) framework~\cite{Fey/Lenssen/2019}, taking the average results of $10$ runs.

\textbf{HGNAS settings.}
We assign $12$ positions for the GNN supernet to cover DGCNN architectures. 
During the design space exploration, the max iteration is set to $1000$, while the population size of $EA$ is set as $20$. 
Searching and training of HGNAS are both conducted on Nvidia V100 GPU. 
During function search and operation search, the number of GNN supernet training epochs is set as $50$ and $500$, respectively.

\textbf{Predictor settings.} 
The predictor is trained for $250$ epochs on $30$K randomly sampled architectures ($21$K for training and $9$K for validation) in our fine-grained design space, with labels obtained from measurement results on various edge devices.
Mean absolute percentage error (MAPE) was used as the loss function during training. 

\textbf{Edge devices.}
We employ four edge devices for comparing HGNAS and competitors: (1) Nvidia RTX3080, (2) Intel i7-8700K, (3) Jetson TX2 with $8$GB memory, (4) Raspberry Pi 3B+ with a Cortex-A5 processor and $1$GB memory. The latency and peak memory usage are obtained by deploying the GNN models on the above devices for inference using the PyG framework.

\subsection{Exploration by HGNAS}

The models designed by HGNAS are comprehensively compared with baselines in terms of model size, accuracy, latency, and peak memory. 

\subsubsection{Accuracy vs. Latency}

\begin{figure}[t]
    \centering
    \includegraphics[width = 0.81\linewidth]{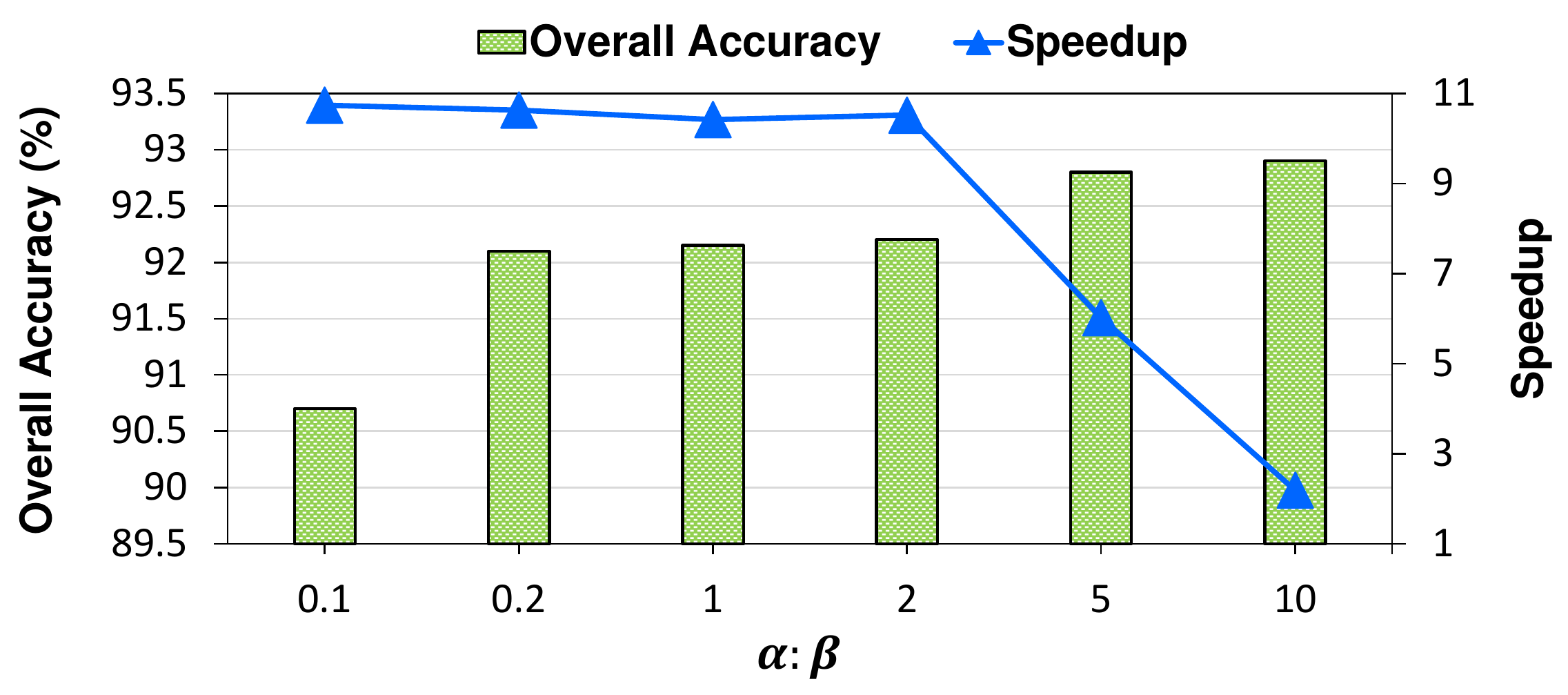}
    \caption{The trade-off between accuracy and speedup (compare to DGCNN) by scaling factor $\alpha$ and $\beta$.}
    \label{fig:tradeoff}
    \vspace{-3pt}
\end{figure}

Fig.~\ref{fig:pareto} reports the exploration results of HGNAS. The ideal solution is located in the top-left corner, indicated by a star. 
The green points named Device\_Acc (e.g. RTX\_Acc) are the optimal architectures designed for the targeted device by HGNAS with no loss of accuracy, while the red points named Device\_Fast allow $1\%$ accuracy loss.
The results show that HGNAS consistently maintains a better performance frontier (higher accuracy and lower latency) on various devices, which is guaranteed by the accurate hardware performance prediction of the candidate architecture during the search.
By setting the scaling factors, HGNAS can easily achieve the tradeoff between hardware efficiency and task accuracy.
Specifically, when $\alpha / \beta$ is smaller, the search results are more in favor of lower latency than higher accuracy.
Conversely, when $\alpha / \beta$ is larger, the search results tend to emphasize more on accuracy.

\begin{table}[t]
  \centering
  \renewcommand\arraystretch{1.1}
  \caption{Comparison of HGNAS and existing models, where OA and mAcc denote overall and balanced accuracy, respectively.}
  \resizebox{\linewidth}{!}{
    \begin{tabular}{c|c|c|c|c|c|c}
    \hline
    Device & Network &Size [MB] & OA   & mAcc & Latency [ms] & Mem. [MB]   \\
    \hline
    \multirow{5}[2]{*}{RTX3080} & DGCNN & 1.81 & 92.9 & 88.9 & 51.8 & 144.0   \\
         & \cite{li2021towards}& - &   92.6   &  89.6   &  (2.0$\times$$\uparrow$)   &    (51.9\%$\downarrow$)   \\
         & \cite{tailor2021towards}& -  &  93.2    &    90.6  & (2.5$\times$$\uparrow$)     &    -  \\
         & RTX-Acc & 1.61 & 92.8 & 90.1 & 8.6 (6.0$\times$$\uparrow$) &   19.1 (86.7\%$\downarrow$)   \\
         & \textbf{RTX-Fast} &  \textbf{1.50} & \textbf{92.1} & \textbf{88.5} & \textbf{4.9 (10.6$\times$$\uparrow$)}  &   \textbf{17.1 (88.1\%$\downarrow$)}  \\
    \hline
    \multirow{5}[2]{*}{i7-8700K} & DGCNN &  1.81 & 92.9 & 88.9 & 234.2 & 643.0      \\
         & \cite{li2021towards} & 2.33&   92.6   &  89.6   &  217.4 (1.1$\times$$\uparrow$)   &   581.3 (9.6\%$\downarrow$)   \\
         & \cite{tailor2021towards} & 1.80 &   93.2   &    90.6  &  92.4 (2.5$\times$$\uparrow$)   &   454.6 (29.3\%$\downarrow$)   \\
         & Intel-Acc  & 1.60 &   92.8   &  89.3    &   77.6 (3.0$\times$$\uparrow$)  & 423.7 (34.1\%$\downarrow$)     \\
         & \textbf{Intel-Fast} & \textbf{1.50} &   \textbf{92.5}   &   \textbf{88.8}   &    \textbf{23.1 (10.2$\times$$\uparrow$)}  &  \textbf{439.2 (31.6\%$\downarrow$)}    \\
    \hline
    \multirow{5}[2]{*}{Jetson TX2} & DGCNN & 1.81  & 92.9 & 88.9 & 270.4 &   145.0    \\
         & \cite{li2021towards} & 2.33 &   92.6   &  89.6   &  109.9 (2.5$\times$$\uparrow$)   &   36.2 (75.2\%$\downarrow$)   \\
         & \cite{tailor2021towards} & 1.81 &  93.2    &    90.6  & 206.4 (1.3$\times$$\uparrow$)    & 29.5 (79.7\%$\downarrow$)      \\
         & TX2-Acc & 1.60 & 92.9 & 89.7 & 50.5 (5.3$\times$$\uparrow$) &   19.0 (86.9\%$\downarrow$)    \\
         & \textbf{TX2-Fast} &  \textbf{1.48} & \textbf{92.2} & \textbf{88.7} & \textbf{36.3 (7.5$\times$$\uparrow$)} &   \textbf{17.1 (88.2\%$\downarrow$)}   \\
    \hline
    \multirow{5}[2]{*}{Raspberry Pi}& DGCNN & 1.81  & 92.9 & 88.9 & 4139.1 &   457.8    \\
         & \cite{li2021towards} & 2.33 &   92.6   &  89.6   &  3466.0 (1.2$\times$$\uparrow$)   &   354.3 (22.6\%$\downarrow$)   \\
         & \cite{tailor2021towards} & 1.81 &  93.2    &    90.6  & 1961.7 (2.1$\times$$\uparrow$)      &   271.1 (40.8\%$\downarrow$)    \\
         & Pi-Acc & 1.50 &   92.8   &   89.3   &  1165.3 (3.6$\times$$\uparrow$)    &   270.2 (41.0\%$\downarrow$)   \\
         & \textbf{Pi-Fast} & \textbf{1.36} &    \textbf{92.1}  &    \textbf{88.3} &   \textbf{557.6 (7.4$\times$$\uparrow$)}  &  \textbf{257.8 (43.7\%$\downarrow$)}  \\
    \hline
    \end{tabular}
  \label{tab:total}
  }
  \vspace{-3pt}
\end{table}

\subsubsection{HGNAS over Existing Graph Neural Architectures}

As shown in Tab.~\ref{tab:total}, compared with the baselines, GNN models designed by HGNAS have better hardware efficiency across all edge computing platforms with similar accuracy.
Such remarkable results are due to the hardware awareness incorporated during exploration.
Compared to DGCNN, HGNAS achieves up to $10.6\times$, $10.2\times$, $7.5\times$ and $7.4\times$ speedup, while reducing $88.1\%$, $31.6\%$, $88.2\%$, $43.7\%$ peak memory usage across four devices.
Moreover, HGNAS on the resource-constrained Jetson TX2 attains the same hardware efficiency as DGCNN on the high-performance Nvidia RTX3080 platform, and reduces peak memory usage by $86.8\%$.
The above results clearly demonstrate that flexibility offered by the fine-grained design space in HGNAS enables the pursuit of exceptional GNN computation efficiency on edge.
For a fairer comparison with manual optimizations in~\cite{li2021towards,tailor2021towards}, we adopt their reported accuracy, inference speedup, and memory reduction as the baseline on the GPU platform.
For other edge platforms, we reproduce these baselines based on PyG, due to the lack of pre-trained models and evaluation results.
These results show that GNN models designed by HGNAS have outperformed the manual optimizations across all platforms, benefiting from the accurate prediction of hardware performance during model explorations.

\subsection{Evaluation on GNN Predictor}

\begin{figure}[t]
    \centering
    \includegraphics[width = 1\linewidth]{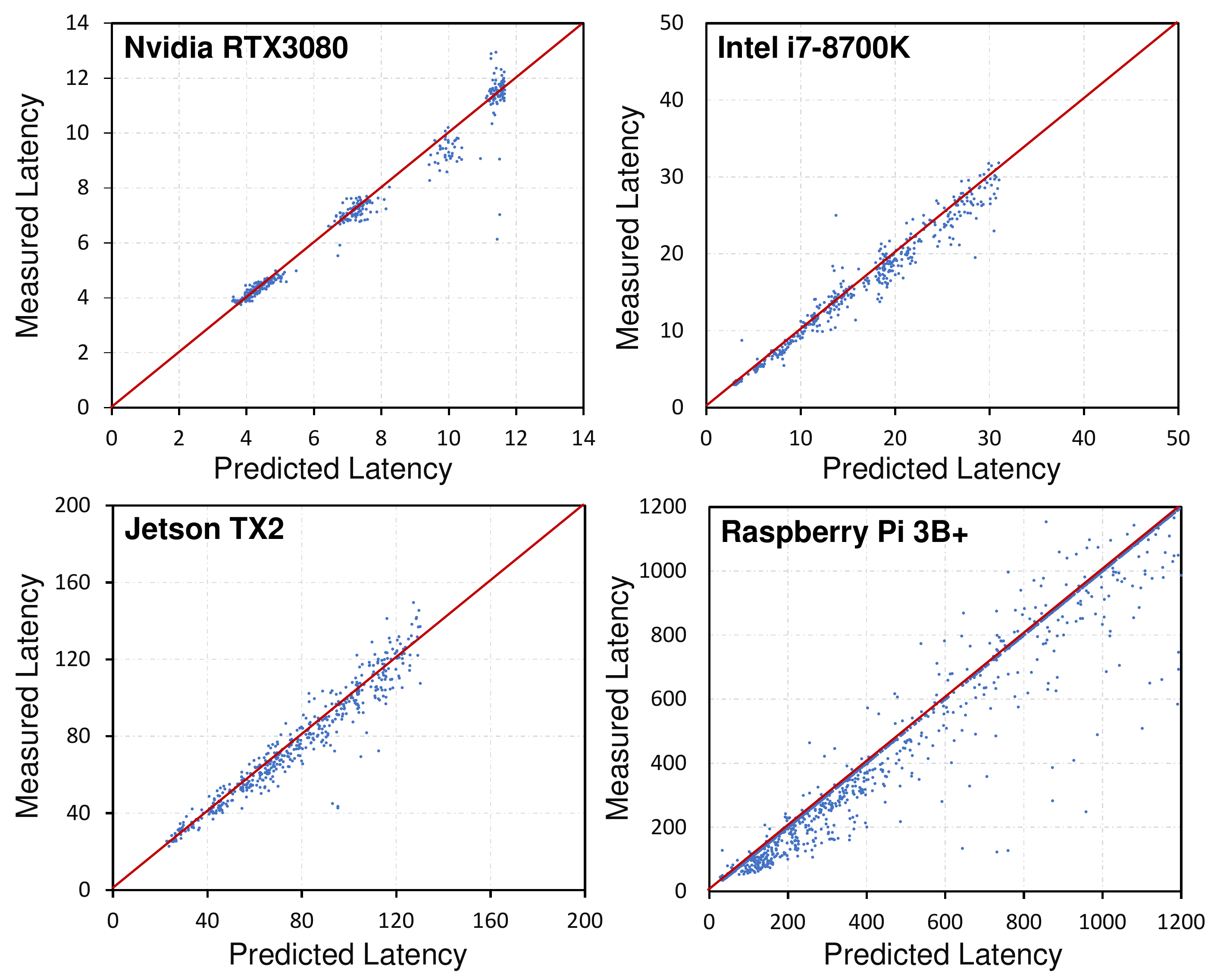}
    \caption{Predictor accuracy on various edge devices. The measured latency denotes the results collected from real edge devices.}
    \label{fig:sactter}
    \vspace{-3pt}
\end{figure}

As shown in Fig.~\ref{fig:sactter}, our proposed hardware performance predictor achieves high prediction accuracy across various platforms.
Specifically, the MAPE of prediction results on RTX3080, Intel i7-8700K, and Jetson TX2 is about $6\%$, while this metric is around $19\%$ on Raspberry Pi due to fluctuations in latency measurement results.
The accuracy of GNN predictor is more than 80\% across devices with a 10\% error bound.
In practice, the GNN predictor obtains better performances for models with a faster inference speed, which assists HGNAS in more efficient design exploration.

\subsection{Ablation Studies}

\begin{figure}[t]
    \centering
    \includegraphics[width = 1\linewidth]{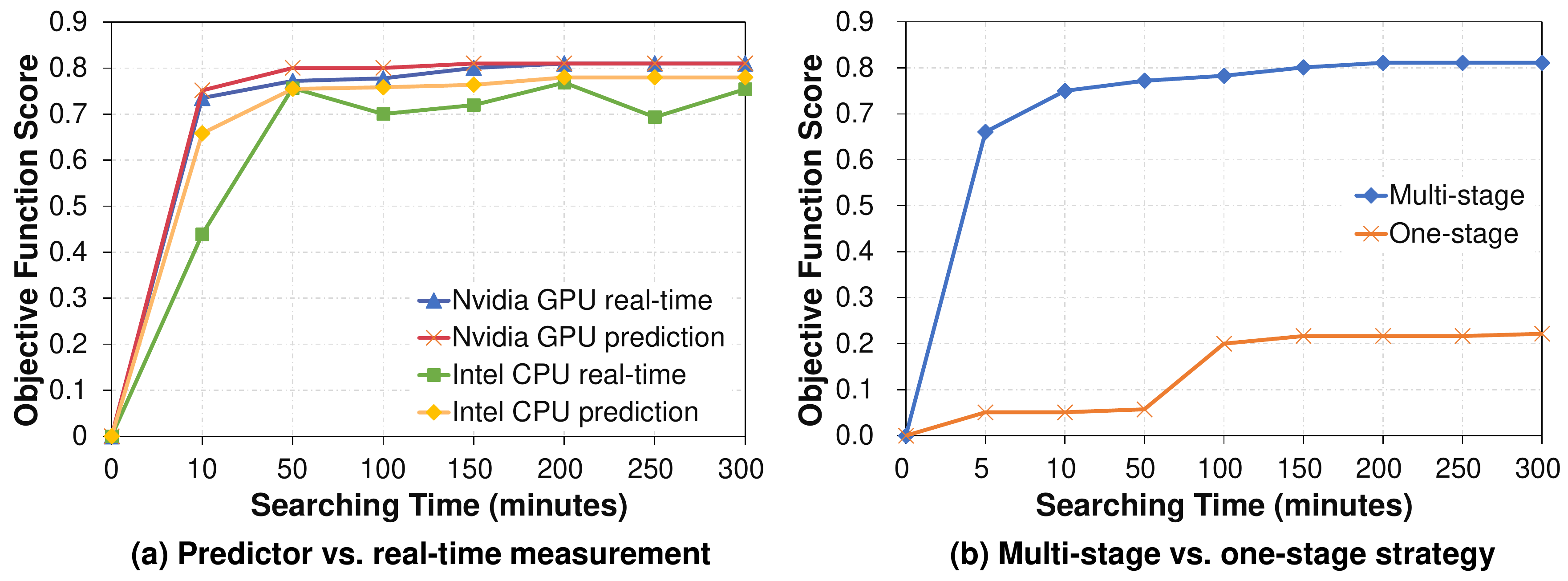}
    \caption{(a) Performance comparison between real-time- and predictor-based search. (b) Search time reduction with the multi-stage strategy.}
    \label{fig:ablation}
    \vspace{-3pt}
\end{figure}

\textbf{Predictor vs. real-time measurement.} 
Fig.~\ref{fig:ablation}(a) shows the HGNAS search process leveraging GNN predictor or real-time measurement on Intel CPU and Nvidia GPU platforms. 
The results demonstrate that the GNN predictor can effectively improve the search efficiency, as the models searched with both methods obtain similar performances.
In particular, our predictor will play a crucial role when the real-time measurement is impossible (e.g., on Jetson TX2 and Raspberry Pi).

\textbf{Multi-stage vs. one-stage search strategy.}
As shown in Fig.~\ref{fig:ablation}(b), exploring with the traditional one-stage search strategy would often be entangled in the huge fine-grained design space.
In contrast, the multi-stage hierarchical search strategy greatly accelerates the exploration process, with the capability of finding an optimal GNN architecture within a few GPU hours.

\subsection{Insight from GNNs Designed by HGNAS}

\begin{figure}[t]
    \centering
    \includegraphics[width = 1\linewidth]{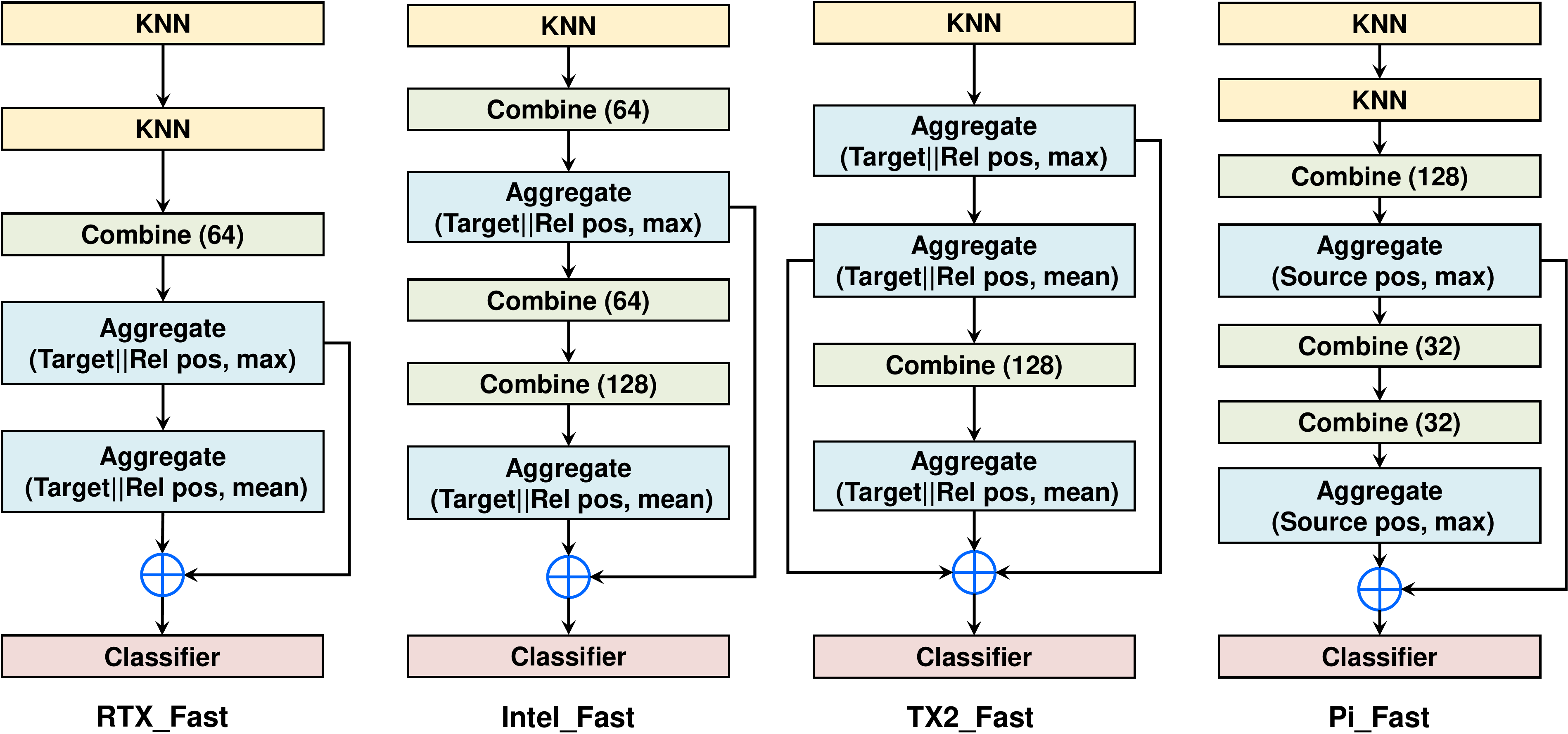}
    \caption{Visualization of GNN models designed by HGNAS.}
    \label{fig:GNNdesigned}
    \vspace{-3pt}
\end{figure}

Fig.~\ref{fig:GNNdesigned} provides the visualization of GNNs designed by HGNAS.
Note that the adjacent KNN operations will be merged during execution due to duplicate graph construction.
The results clearly show that the hardware-efficient architectures designed by HGNAS are closely associated with the characteristics of the target device,
which are consistent with the characterization of GNN models in \textbf{Observation \raisebox{-0.1em}{\ding[1.3]{174}}}. 
For example, as KNN occupies the majority of execution time on RTX3080 and Jetson TX2, GNN models designed for these devices would comprise fewer valid KNN operations.
Moreover, the optimal model for Intel CPU has fewer aggregate operations, and models designed for Raspberry Pi tend to simplify each operation.

%% file: 5-conclusions.tex
\section{Conclusions}

In this paper, we propose HGNAS, the first hardware-aware framework to explore efficient graph neural architecture for edge devices.
HGNAS can automatically search for optimal GNN architectures that maximize both task accuracy and computation efficiency.
HGNAS leverages the multi-stage hierarchical search strategy and GNN hardware performance predictor to efficiently explore the fine-grained GNN design space.
Extensive experiments show that GNN models generated by HGNAS consistently outperform SOTA GNNs, achieving about $10.6\times$ speedup and $88.2\%$ peak memory reduction across various edge platforms.
We believe that HGNAS has made pivotal progress in bringing GNNs to real-life edge applications.

%% file: main.bbl
\begin{thebibliography}{10}

\bibitem{kipf2017semi}
Thomas~N. Kipf and Max Welling.
\newblock Semi-supervised classification with graph convolutional networks.
\newblock In {\em Proceedings of ICLR}, 2017.

\bibitem{zhang2018link}
Muhan Zhang and Yixin Chen.
\newblock Link prediction based on graph neural networks.
\newblock In {\em Proceedings of NIPS}, 2018.

\bibitem{ying2018graph}
Rex Ying, Ruining He, et~al.
\newblock Graph convolutional neural networks for web-scale recommender
  systems.
\newblock In {\em Proceedings of SIGKDD}, 2018.

\bibitem{qi2017pointnet++}
Charles~Ruizhongtai Qi, Li~Yi, et~al.
\newblock {PointNet++}: Deep hierarchical feature learning on point sets in a
  metric space.
\newblock In {\em Proceedings of NIPS}, 2017.

\bibitem{wang2019dynamic}
Yue Wang, Yongbin Sun, et~al.
\newblock Dynamic graph {CNN} for learning on point clouds.
\newblock {\em ACM TOG}, pages 1--12, 2019.

\bibitem{li2021towards}
Yawei Li, He~Chen, et~al.
\newblock Towards efficient graph convolutional networks for point cloud
  handling.
\newblock In {\em Proceedings of ICCV}, 2021.

\bibitem{tailor2021towards}
Shyam~A Tailor, Ren{\'e} de~Jong, et~al.
\newblock Towards efficient point cloud graph neural networks through
  architectural simplification.
\newblock In {\em Proceedings of ICCV}, 2021.

\bibitem{zhang2021exploration}
Jie-Fang Zhang and Zhengya Zhang.
\newblock Exploration of energy-efficient architecture for graph-based
  point-cloud deep learning.
\newblock In {\em Processings of SiPS}, 2021.

\bibitem{dudziak2020brp}
Lukasz Dudziak, Thomas Chau, et~al.
\newblock {BRP-NAS}: Prediction-based {NAS} using {GCNs}.
\newblock In {\em Proceedings of NIPS}, 2020.

\bibitem{wu2019fbnet}
Bichen Wu, Xiaoliang Dai, et~al.
\newblock F{BN}et: Hardware-aware efficient convnet design via differentiable
  neural architecture search.
\newblock In {\em Proceedings of CVPR}, 2019.

\bibitem{ijcai2020p195}
Yang Gao, Hong Yang, et~al.
\newblock Graph neural architecture search.
\newblock In {\em Proceedings of IJCAI}, pages 1403--1409, 2020.

\bibitem{gao2022graphnas++}
Yang Gao, Peng Zhang, et~al.
\newblock Graph{NAS}++: Distributed architecture search for graph neural
  networks.
\newblock {\em IEEE TKDE}, 2022.

\bibitem{wu20153d}
Zhirong Wu, Shuran Song, et~al.
\newblock {3D} shapenets: A deep representation for volumetric shapes.
\newblock In {\em Proceedings of CVPR}, 2015.

\bibitem{cai2021rethinking}
Shaofei Cai, Liang Li, et~al.
\newblock Rethinking graph neural architecture search from message-passing.
\newblock In {\em Proceedings of CVPR}, 2021.

\bibitem{zhang2022pasca}
Wentao Zhang, Yu~Shen, et~al.
\newblock {PaSca}: A graph neural architecture search system under the scalable
  paradigm.
\newblock In {\em Proceedings of WWW}, 2022.

\bibitem{zhang2020autoshrink}
Tunhou Zhang, Hsin-Pai Cheng, et~al.
\newblock {AutoShrink}: A topology-aware {NAS} for discovering efficient neural
  architecture.
\newblock In {\em Proceedings of AAAI}, 2020.

\bibitem{sheng2022larger}
Yi~Sheng, Junhuan Yang, et~al.
\newblock The larger the fairer? small neural networks can achieve fairness for
  edge devices.
\newblock In {\em Proceddings of DAC}, 2022.

\bibitem{profiler}
{PyTorch Profiler}.
\newblock \url{https://pytorch.org/docs/stable/profiler.html}.
\newblock Accessed Nov, 2022.

\bibitem{yan2020characterizing}
Mingyu Yan, Zhaodong Chen, et~al.
\newblock Characterizing and understanding {GCNs} on gpu.
\newblock {\em IEEE CAL}, pages 22--25, 2020.

\bibitem{zhang2021g}
Yongan Zhang, Haoran You, et~al.
\newblock {G-CoS}: {GNN}-accelerator co-search towards both better accuracy and
  efficiency.
\newblock In {\em Proceedings of ICCAD}, 2021.

\bibitem{benmeziane2021comprehensive}
Hadjer Benmeziane, Kaoutar~El Maghraoui, et~al.
\newblock A comprehensive survey on hardware-aware neural architecture search.
\newblock {\em arXiv preprint arXiv:2101.09336}, 2021.

\bibitem{guo2020single}
Zichao Guo, Xiangyu Zhang, et~al.
\newblock Single path one-shot neural architecture search with uniform
  sampling.
\newblock In {\em Proceedings of ECCV}. Springer, 2020.

\bibitem{Fey/Lenssen/2019}
Matthias Fey and Jan~Eric Lenssen.
\newblock Fast graph representation learning with {PyTorch Geometric}.
\newblock In {\em Proceedings of ICLR}, 2019.

\end{thebibliography}
